\documentclass[]{article}
\makeatletter\if@twocolumn\PassOptionsToPackage{switch}{lineno}\else\fi\makeatother

      \makeatletter
\usepackage{wrapfig}
\usepackage{graphics} 
\usepackage{epsfig} 
\usepackage{mathptmx} 
\usepackage{times} 
\usepackage{amsmath} 
\usepackage{amssymb}  
\usepackage{csvsimple}
\usepackage{array}
\usepackage{float}
\usepackage{verbatim}
\usepackage{subfig}
\graphicspath{ {./} }
\usepackage{makecell}
\usepackage{hyperref}
\usepackage{caption}
\newcounter{aubio}
\usepackage{tabularx}
\usepackage{booktabs}
\usepackage{authblk}

\long\def\bioItem{%
\@ifnextchar[{\@bioItem}{\@@bioItem}}

\long\def\@bioItem[#1]#2#3{
 \stepcounter{aubio}
 \expandafter\gdef\csname authorImage\theaubio\endcsname{#1}
 \expandafter\gdef\csname Rokas Jurevičius\theaubio\endcsname{#2}
 \expandafter\gdef\csname authorDetails\theaubio\endcsname{#3}
}

\long\def\@@bioItem#1#2{
 \stepcounter{aubio}
 \expandafter\gdef\csname authorName\theaubio\endcsname{#1}
 \expandafter\gdef\csname authorDetails\theaubio\endcsname{#2}
}

\newcommand{\checkheight}[1]{%
  \par \penalty-100\begingroup%
  \setbox8=\hbox{#1}%
  \setlength{\dimen@}{\ht8}%
  \dimen@ii\pagegoal \advance\dimen@ii-\pagetotal
  \ifdim \dimen@>\dimen@ii
    \break
  \fi\endgroup}

\def\printBio{%
  \@tempcnta=0
   \loop
     \advance \@tempcnta by 1
     \def\aubioCnt{\the\@tempcnta}
     \setlength{\intextsep}{0pt}%
     \setlength{\columnsep}{10pt}%
     \expandafter\ifx\csname authorImage\aubioCnt\endcsname\relax%
      \else%
       \checkheight{\includegraphics[height=1.25in,width=1in,keepaspectratio]{\csname authorImage\aubioCnt\endcsname}}
        \begin{wrapfigure}{l}{25mm}
         \includegraphics[height=1.25in,width=1in,keepaspectratio]{\csname authorImage\aubioCnt\endcsname}
        \end{wrapfigure}\par
      \fi
     \noindent\textbf{\csname authorName\aubioCnt\endcsname}\csname authorDetails\aubioCnt\endcsname \par\bigskip
      \ifnum\@tempcnta < \theaubio
   \repeat
   }
\makeatother

\usepackage{amsmath,tabulary,graphicx,times,caption,fancyhdr,amssymb,amsfonts,amstext,amsbsy}
\usepackage[utf8]{inputenc}
\usepackage[paperheight=10in,paperwidth=6.5in,margin=2cm,headsep=.5cm,top=2.5cm]{geometry}
\renewenvironment{abstract} {\vspace*{-1pc}\trivlist\item[]\leftskip\oupIndent\hrulefill\par\vskip4pt\noindent\textbf{\abstractname}\mbox{\null}\\}{\par\noindent\hrulefill\endtrivlist} 
 \date{}
\usepackage{url,multirow,morefloats,floatflt,cancel,tfrupee}
\makeatletter

\AtBeginDocument{\@ifpackageloaded{textcomp}{}{\usepackage{textcomp}}}
\makeatother
\usepackage{colortbl}
\usepackage{xcolor}
\usepackage{pifont}
\usepackage[nointegrals]{wasysym}
\makeatletter

\def\mcWidth#1{\csname TY@F#1\endcsname+\tabcolsep}

\def\cAlignHack{\rightskip\@flushglue\leftskip\@flushglue\parindent\z@\parfillskip\z@skip}
\def\rAlignHack{\rightskip\z@skip\leftskip\@flushglue \parindent\z@\parfillskip\z@skip}

\usepackage{ifxetex}
\ifxetex\else\if@twocolumn\usepackage{dblfloatfix}\fi\fi

\AtBeginDocument{
\expandafter\ifx\csname eqalign\endcsname\relax
\def\eqalign#1{\null\vcenter{\def\\{\cr}\openup\jot\m@th
  \ialign{\strut$\displaystyle{##}$\hfil&$\displaystyle{{}##}$\hfil
      \crcr#1\crcr}}\,}
\fi
}

\AtBeginDocument{%
  \@ifpackageloaded{endfloat}%
   {\renewcommand\efloat@iwrite[1]{\immediate\expandafter\protected@write\csname efloat@post#1\endcsname{}}}{\newif\ifefloat@tables}%
}%

\def\BreakURLText#1{\@tfor\brk@tempa:=#1\do{\brk@tempa\hskip0pt}}
\let\lt=<
\let\gt=>
\def\processVert{\ifmmode|\else\textbar\fi}

\@ifundefined{subparagraph}{
\def\subparagraph{\@startsection{paragraph}{5}{2\parindent}{0ex plus 0.1ex minus 0.1ex}%
{0ex}{\normalfont\small\itshape}}%
}{}

\newcommand\role[1]{\unskip}
\newcommand\aucollab[1]{\unskip}
  
\@ifundefined{tsGraphicsScaleX}{\gdef\tsGraphicsScaleX{1}}{}
\@ifundefined{tsGraphicsScaleY}{\gdef\tsGraphicsScaleY{.9}}{}
\def\checkGraphicsWidth{\ifdim\Gin@nat@width>\linewidth
	\tsGraphicsScaleX\linewidth\else\Gin@nat@width\fi}

\def\checkGraphicsHeight{\ifdim\Gin@nat@height>.9\textheight
	\tsGraphicsScaleY\textheight\else\Gin@nat@height\fi}

\def\fixFloatSize#1{}
\let\ts@includegraphics\includegraphics

\def\inlinegraphic[#1]#2{{\edef\@tempa{#1}\edef\baseline@shift{\ifx\@tempa\@empty0\else#1\fi}\edef\tempZ{\the\numexpr(\numexpr(\baseline@shift*\f@size/100))}\protect\raisebox{\tempZ pt}{\ts@includegraphics{#2}}}}

\AtBeginDocument{\def\includegraphics{\@ifnextchar[{\ts@includegraphics}{\ts@includegraphics[width=\checkGraphicsWidth,height=\checkGraphicsHeight,keepaspectratio]}}}

\DeclareMathAlphabet{\mathpzc}{OT1}{pzc}{m}{it}

\def\URL#1#2{\@ifundefined{href}{#2}{\href{#1}{#2}}}

\def\UrlOrds{\do\*\do\-\do\~\do\'\do\"\do\-}%
\g@addto@macro{\UrlBreaks}{\UrlOrds}

\@ifundefined{quoteAttrib}
	{}
	{}

\@ifundefined{titlequoteAttrib}
	{}{}

\newenvironment{title-quote}
	{\list{}{\fontsize{10pt}{12pt}\selectfont\leftmargin.5in\itshape\rightmargin\leftmargin}%
  \item\relax}
  {\endlist}

\makeatother

\usepackage[noindentafter]{titlesec}
\def\NormalBaseline{\def\baselinestretch{1.1}}

\titleformat{\section}[hang]{\NormalBaseline\filright\large\fontsize{12}{15}\bfseries\boldmath}
{\large\thesection.}
{10pt}
{\noindent}
[]
\titleformat{\subsection}[hang]{\NormalBaseline\filright\fontsize{11}{13}\bfseries\itshape\boldmath}
{\thesubsection.}
{10pt}
{}
[]
\titleformat{\subsubsection}[hang]{\NormalBaseline\filright\fontsize{10}{12}\bfseries\itshape\boldmath}
{\thesubsubsection.}
{10pt}
{}
[]
\titleformat{\paragraph}[runin]{\NormalBaseline\filright\itshape}
{\theparagraph.}
{10pt}
{}
[]
\titleformat{\subparagraph}[runin]{\NormalBaseline\filright\itshape}
{\thesubparagraph.}
{10pt}
{}
[]

\titlespacing{\section}{0pt}{1.5\baselineskip}{.2\baselineskip}  
\titlespacing{\subsection}{0pt}{1\baselineskip}{.2\baselineskip}  
\titlespacing{\subsubsection}{0pt}{1.5\baselineskip}{.2\baselineskip}  
\titlespacing{\paragraph}{0pt}{.5\baselineskip}{10pt}  
\titlespacing{\subparagraph}{0pt}{.5\baselineskip}{10pt}

\makeatletter\def\oupIndent{1pt}
\def\author#1{\gdef\@author{\hskip-\dimexpr(\tabcolsep)\hskip\oupIndent\parbox{\dimexpr\textwidth-\oupIndent}{\centering\bfseries#1}}}
\def\title#1{\gdef\@title{\centering\bfseries\ifx\@articleType\@empty\else\@articleType\\\fi#1}}
\let\@articleType\@empty \def\articletype#1{\gdef\@articleType{{\normalfont\itshape#1}}}
\fancypagestyle{headings}{\fancyhf{}\fancyhead[C]{\RunningHead}\fancyhead[R]{\thepage}}
\makeatother
\usepackage[authoryear]{natbib}

\begin{document}

\author{Rokas Jurevi\v{c}ius\textsuperscript{1}\thanks{Corresponding author. E-mail: rokas.jurevicius@mii.vu.lt}~ and
            Virginijus Marcinkevi\v{c}ius\textsuperscript{1}\thanks{E-mail: virginijus.marcinkevicius@mii.vu.lt}{ }~\\[-3pt]\normalsize\normalfont\itshape 
~\\\textsuperscript{1}{Institute of Data Science and Digital Technologies\unskip, Vilnius University\unskip, Akademijos g. 4\unskip, Vilnius\unskip, Lithuania}}
\title{AIR: A Dataset of Aerial Imagery from Robotics Simulator for Map-Based Localization Systems Benchmark}

\def\RunningHead{{AIR: A Dataset of Aerial Imagery from Robotics Simulator for Map-Based Localization Systems Benchmark}}

\maketitle

\begin{abstract}

\textbf{Purpose} -- This paper presents a new dataset of Aerial Imagery from Robotics simulator (abbr. AIR). AIR dataset aims to provide a starting point for localization system development and to become a typical benchmark for accuracy comparison of map-based localization algorithms, visual odometry, and SLAM for high altitude flights.\\

\textbf{Design/methodology/approach} -- The presented dataset contains over 100 thousand aerial images captured from Gazebo robotics simulator using orthophoto maps as a ground plane. Flights with 3 different trajectories are performed on maps from urban and forest environment at different altitudes, totaling over 33 kilometers of flight distance. \\

\textbf{Findings} -- The review of previous researches show, that the presented dataset is the largest currently available public dataset with downward facing camera imagery. \\

\textbf{Originality/value} -- This paper presents the problem of missing publicly available datasets for high altitude (100--3000 meters) UAV flights, the current state-of-the-art researches performed to develop map-based localization system for UAVs, depend on real-life test flights and custom simulated datasets for accuracy evaluation of the algorithms. The presented new dataset solves this problem and aims to help the researchers to improve and benchmark new algorithms for high-altitude flights.

\end{abstract}


\section{Introduction}

The problem of GPS-denied navigation for unmanned aerial vehicles (abbr. UAV) is getting more attention with the increasing popularity of consumer and commercial UAVs. The problem of GPS-denied arises due to the weakness of GPS systems that they rely on radio signals which can be jammed or spoofed. In case the radio signals are jammed during a flight, there are only a limited number of technologies that can localize the aircraft. Any radio aided localization system is unusable if radio signals are jammed. It is possible to use Inertial Navigation Systems, but they are very expensive. Therefore, less expensive Visual localization methods with an optical camera sensor can be used to localize the aircraft. A number of methods have been proposed to solve GPS-denied localization problem using imaging sensors and computer vision algorithms. A survey by \cite{lu2018survey} suggests, that current methods can be categorized into map-based, map-building, and map-less systems. Map-less systems include visual odometry and optical flow algorithms. Map-building systems are usually recognized as Simultaneous Localization And Mapping (abbr. SLAM) algorithms, that build maps of the environment during runtime and localizes relatively to the created map. Both map-less and map-building systems have a number of public datasets that are widely used by the computer vision community to benchmark these systems. Map-based systems do not have established public datasets and researchers create their own datasets to measure their accuracy. This problem was already mentioned in research papers by \cite{nassar2018deep} and \cite{shan2015google}. The self-made datasets are usually small since a creation of a dataset is very laborious. Due to the absence of publicly available datasets, the comparison between algorithms becomes ambiguous. An initial off-line benchmarking can provide a substantial information about the algorithm performance within common pitfalls and can be used to compare against each other while leaving the real-life test flights for the last stages of development. This paper presents a new dataset that was specifically created for map-based system benchmark providing different trajectories, environments, and altitudes. Despite a map-based system focus, it can be also used for map-less and map-building systems to measure the accuracy for high altitude (100--3000 meters) UAV localization. Publisher node for robot operating system (abbr. ROS) is also available to easily start using the dataset in the ROS environment.

\begin{figure}
\centering
\includegraphics[scale=0.28]{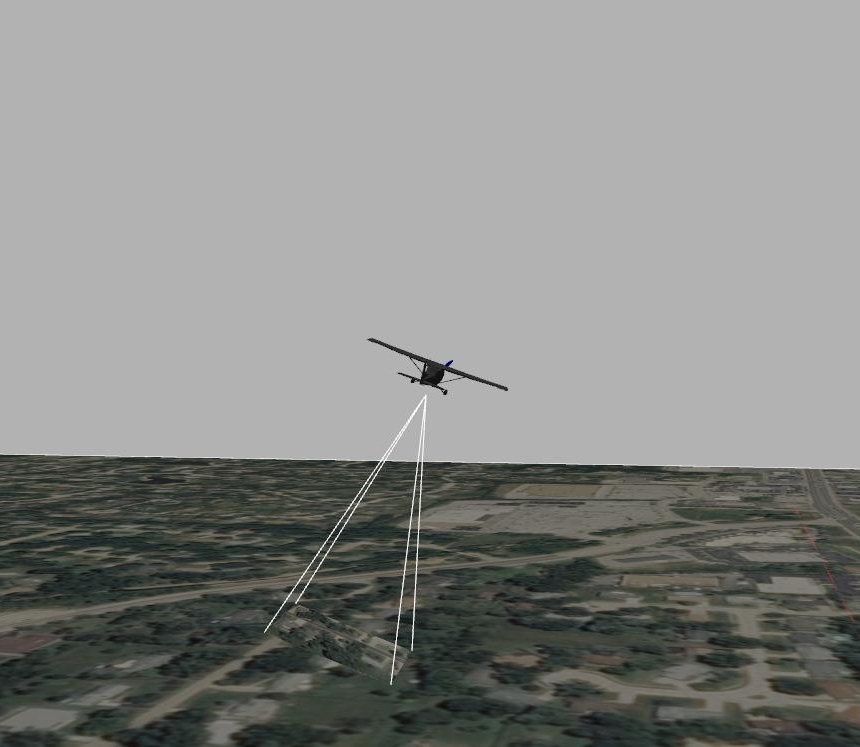}
\caption{A visualization of a plane model flying over an urban environment.}
\label{fig:simulator}
\end{figure}

\section{Aerial Imagery Datasets} \label{sec:datasets}

This section reviews popular datasets used for aerial localization systems benchmarks. Since there is no public dataset suitable for map-based system benchmark, recent works of map-building localization systems are also reviewed, to observe the volume of data used in the state-of-the-art map-based localization system development.

\subsection{Most notable publicly available datasets for the benchmark of localization algorithms}
\label{sec:existing_datasets}

A few datasets that have been established as a baseline for benchmarking of SLAM and Visual Odometry algorithms accuracy. The most notable are KITTI \cite{geiger2013vision}, TUM-RGBD \cite{sturm2012benchmark}, ICL-NUIM \cite{handa2014benchmark}, EUROC \cite{burri2016euroc} and a very recent Zurich Urban Micro Aerial Vehicle Dataset \cite{majdik2017zurich}. The following list provides short content overviews of these datasets:
\begin{enumerate}
\item The TUM-RGBD dataset was created using a hand-held Kinect sensor in an indoors environment. The dataset provides ground truth locations of the camera that were precisely captured, using external sensors mounted in the environment.
\item The ICL-NUIM dataset provides hand-held RGB-D data sequences from a simulated environment of small indoor spaces, such as living room and office. Ground truth contains camera locations, the dataset is very similar to TUM-RGBD, except it was created in a simulated environment.
\item The EUROC dataset is created in various indoors environments with a stereo camera mounted on a micro UAV. The dataset contains precise ground truth trajectory of the aircraft from VICON motion capture system and 3D scans of the environments.
\item The KITTI dataset was created using a stereo camera and other sensors mounted on a car and the imagery was captured while driving in an outdoor urban environment. The dataset contains a lot of material (130 gigabytes of video footage) and can be used as a benchmark for localization algorithms that use a front facing camera images for UAVs flying at low altitude.
\item The Zurich dataset was specifically created to benchmark localization systems of UAVs flying in low altitude in outdoor urban environments. The Zurich dataset contains monocular front-facing camera images captured from a UAV flying at the altitude of 5-15 meters. The dataset contains images from 2 kilometers of flight distance in Zurich city. 
\end{enumerate}

\subsection{Datasets used for map-based systems benchmarking}

The dataset reviewed in section \ref{sec:existing_datasets} provides a variety of challenging environments --- indoor and industrial spaces, urban city streets with live traffic. But there is no dataset that could be used to measure localization accuracy of a high altitude UAV flight with a downward facing camera. Due to the lack of a dataset, the following list of researches relies on self-made simulated datasets or real-life test flights:
\begin{enumerate}
\item Navigation system using visual odometry, inertial navigation, and image registration proposed by \cite{conte2009vision} uses images from an off-line dataset which contains imagery from a 1-kilometer distance flight of a rectangular trajectory. The proposed system was additionally evaluated during a test flight.
\item Geo-referenced localization approach published by \cite{lindsten2010geo} was evaluated using a self-made dataset captured from a flight covering a distance of 400 meters.
\item Map matching approach presented by \cite{shan2015google} was evaluated using images from a 3-minute flight (around 360 meters of distance) at an altitude of 80 meters. This dataset is available publicly and can be used for benchmarking, although it is rather small and contains only a single flight.
\item A research by \cite{nassar2018deep} proposed deep Convolutional Neural Network (abbr. CNN) for image registration from a high altitude UAV flight. Due to the lack of datasets, authors created two datasets of their own, (1) using images from 1.2-kilometer distance flight at 300 meters altitude over the city of Potsdam, Germany and (2) from a flight of 0.5-kilometer distance over Famagusta, Cyprus. 
\item Navigation system based on detection and matching of road intersections proposed by \cite{dumble2015airborne} evaluates the system on a test flight of around 7 kilometers distance. 
\item  Image feature based localization approach is proposed by \cite{wang2013vision} and the system is evaluated using a self-made image-in-loop simulation. The volume of the dataset is undisclosed.
\end{enumerate}
   
Unfortunately, datasets from most of these researches are not publicly available or the volume of the datasets are relatively small and does not cover more than one aspect of map-based GPS-denied localization.

\section{AIR: A New Dataset for Benchmark of Map-Based Localization Algorithms}

Section 2 has established the problem of missing publicly available dataset of high altitude UAV flight. This section draws requirements for the new dataset, describes the simulation environment where flights were performed, and provides details of the dataset contents.

\begin{figure*}
\centering
\includegraphics[width=\textwidth]{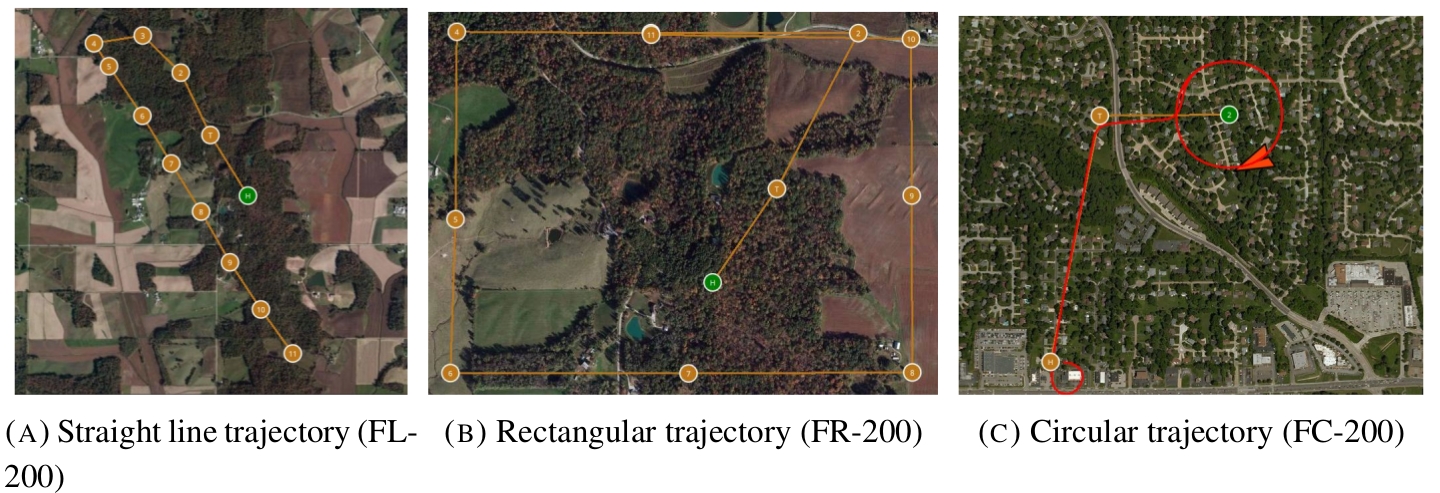}
\caption{Flight plans for different scenarios.}
\label{fig:trajectories}
\end{figure*}

\subsection{Dataset Requirements}
\label{sec:requirements}

To define requirements for the new dataset, an overview of common map-based localization system components and their pitfalls are needed. Map-based localization systems use image registration algorithms to match the image from an on-board camera with a map. These algorithms are prone to errors when the maps are not up-to-date, this problem is presented in detail in section \ref{sec:robustness}. Another common component is visual odometry, which provides motion information between the images. Usually, visual odometry relies on a high framerate video feed to accurately calculate vehicle motion in real-time. For the dataset to be suitable for visual odometry, authors of popular visual odometry algorithms suggest, that the image sequences should be of at least 50 frames per second \cite{forster2014svo}, \cite{klein2007parallel}. From the researches reviewed in section \ref{sec:datasets}, it is clear that researchers tend to use imagery from urban environments. Since urban environments contain buildings, roads, and other artifacts, such imagery provides rich texture that computer vision algorithms can successfully process and extract information. An additional environment, such as forests, would provide useful information about system accuracy, when the imagery is not rich with texture and features. Different trajectories at different altitudes should also provide additional validation of system behavior in different flight scenarios. The creation of a dataset is very laborious, so researchers might be using small datasets to evaluate localization accuracy which might not expose the long-term issues of localization drift, so the distance should also be considered as a key requirement for the dataset. Advanced localization systems are using loop-closure algorithms to improve localization accuracy and reduce drift over time, these algorithms detect a point in space that was already passed during the localization and optimizes the flight trajectory to match the passed point. It would be beneficial for a system to test this feature using the new dataset. After the careful review of localization systems, the key requirements for the dataset were defined as follows:
\begin{enumerate}
\item flights should be performed in multiple environments: urban and forest,
\item image capture framerate should be at least 50 frames per second,
\item maps of different dates should be available to test algorithm robustness to changes in the imagery,
\item different trajectories should be available,
\item flight distances should be 1 kilometer or longer.
\end{enumerate}

\begin{figure*}
\centering
\includegraphics[width=0.6\textwidth]{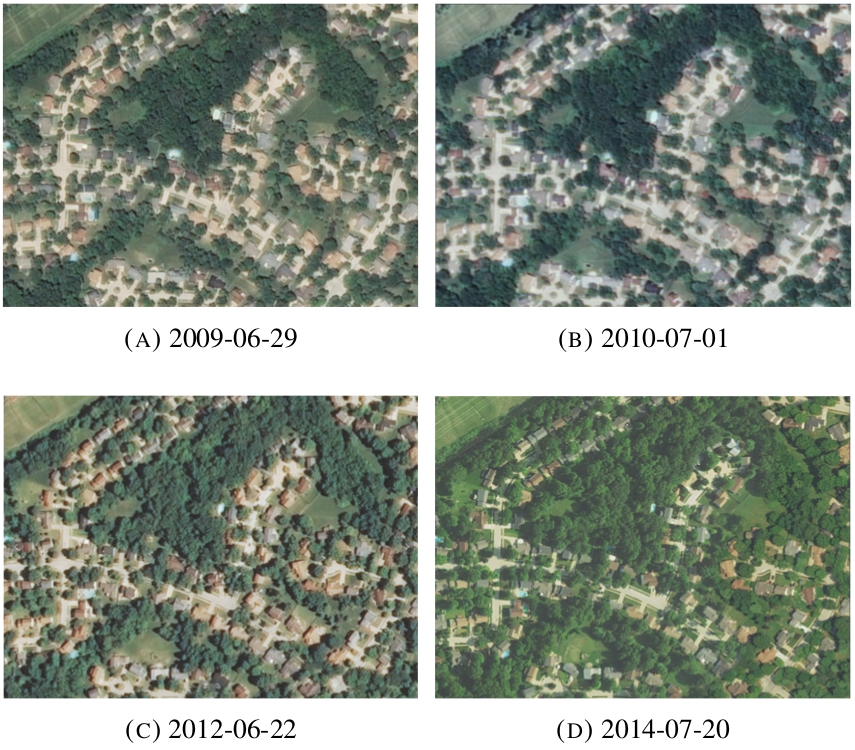}
\caption{Sample patches of the same location in urban map created at different dates.}
\label{fig:map_robustness}
\end{figure*}

\subsection{Flight Scenarios}

\label{sec:scenarios}
A number of different flight scenarios are set up to satisfy requirements defined in section \ref{sec:requirements}. A single flight scenario is a combination of a map, trajectory, and altitude. Table \ref{tab:scenarios} shows different trajectories, maps, and altitudes that are used while performing simulated flights. 
Three trajectories were chosen:
\begin{enumerate}
\item straight line
\item rectangular
\item circular
\end{enumerate}
Previews of the trajectories are given in figure \ref{fig:trajectories}. Rectangular and circular trajectories have a returning point, so these scenarios can be used to test loop-closure algorithms. Each flight scenario is given an abbreviation, e.g. FL-200, where the first character stands for a map (F - forest, U - urban), the second character stands for trajectory (L - straight line, R - rectangular, C - circular), and the number stands for altitude in meters. A total of 12 flight scenarios is performed.  

\begin{table}[H]
    \caption{Different trajectories, maps, and altitudes used to perform 12 flight scenarios in the dataset\label{tab:scenarios}}
    \centering
\begin{tabular} {l l l}%
    \bfseries Trajectory 
    & \bfseries Map
    & \bfseries Altitude, meters
    \begin{filecontents*}{scenarios.csv}
Trajectory,Map,Altitude meters
Line (L),Forest (F),200
Rectangular (R),Urban (U),300
Circular (C),,
\end{filecontents*}
\csvreader[head to column names]{scenarios.csv}{}
    {\\\hline
       \csvcoli   & 
       \csvcolii  &
       \csvcoliii
       }
\end{tabular}
\end{table}

\subsection{Simulation Environment}

Dataset was created using orthophoto maps from the United States Geological Survey's (abbr. USGS) National Agriculture Imagery Program (abbr. NAIP) database \cite{usdaoffice} and Gazebo Robotics simulation environment. Maps were used as a texture for the simulator ground plane, a preview of a plane model flying in the simulator environment is shown in figure \ref{fig:simulator}. Flights over urban environment used the map created at the year 2009 and flights over forest environment used the map created in 2008. PX4 flight controller software in software in the loop (abbr. SITL) mode was used to control the aircraft model in the simulated environment. A software service was developed which captured images from the simulator as soon as it reached target altitude and location. Image meta data was collected using the CSV file format. Flight trajectories for each scenario were planned manually using QGroundControl\footnote{Qgroundcontrol: Ground control station for small air land water autonomous unmanned systems, website: \url{http://qgroundcontrol.com/}} software. The flight plans were also saved alongside the images. Previews of planned trajectories from the ground control software are shown in figure \ref{fig:trajectories}. The flight plans includes plane takeoff, but this part is not included in the dataset, and image sequences begin when the plane reaches it's starting point and altitude.

\subsection{Creation of the dataset}

The dataset is created by capturing images and sensor data during the simulated flight in the Gazebo simulation environment. A custom software service was used to subscribe to images and sensor data, recording the images in JPEG format and writing other sensor data into a CSV file. The images are generated from a simulated camera sensor attached to the bottom of the aircraft model. The sensor is facing downward, and it has $ 57^\circ $ field of view, white lines in figure \ref{fig:simulator} depict the field of view of the camera. The camera is attached to the frame of the model rigidly, so it captures all perspective distortions caused by the maneuvers of the aircraft. Images are captured at 50 Hz frequency during the execution of the flight plan. Takeoff is not captured, the capture starts when the cruise altitude (200 or 300 meters, depending on the flight plan) is reached.

\subsection{Robustness to inaccuracies of map imagery}
\label{sec:robustness}

The dataset is created by capturing images and sensor data during the simulated flight in the Gazebo simulation environment. A custom software service was used to subscribe to images and sensor data, recording the images in JPEG format and writing other sensor data into a CSV file. The images are generated from a simulated camera sensor attached to the bottom of the aircraft model. The sensor is facing downward, and it has $ 57^\circ $ field of view, white lines in figure \ref{fig:simulator} depict the field of view of the camera. The camera is attached to the frame of the model rigidly, so it captures all perspective distortions caused by the maneuvers of the aircraft. Images are captured at 50 Hz frequency during the execution of the flight plan. \\ Takeoff is not captured, the capture starts when the cruise altitude (200 or 300 meters, depending on the flight plan) is reached.

\begin{table}[H]
    \caption{Number of images and distances of each flight.\label{tab:flight_stats}}
    \centering
\begin{tabular} {l l l}%
    \bfseries Scenario 
    & \bfseries Image count
    & \bfseries Flight distance, meters
    \begin{filecontents*}{stats.csv}
Scenario,Number of images,Flight distance
FL-200,11837,2936
FL-300,9020,2736
FR-200,16056,4748
FR-300,15307,4541
FC-200,3601,1065
FC-300,3450,1028
UL-200,8735,2592
UL-300,8416,2496
UR-200,14997,4418
UR-300,15361,4542
UC-200,3422,1016
UC-300,3272,970
Total,113474,33088
\end{filecontents*}
\csvreader[head to column names]{stats.csv}{}
    {\\\hline
       \csvcoli   & 
       \csvcolii  &
       \csvcoliii
       }
\end{tabular}
\end{table}

\subsection{Dataset contents}

\begin{figure*}
\centering
\includegraphics[width=\textwidth]{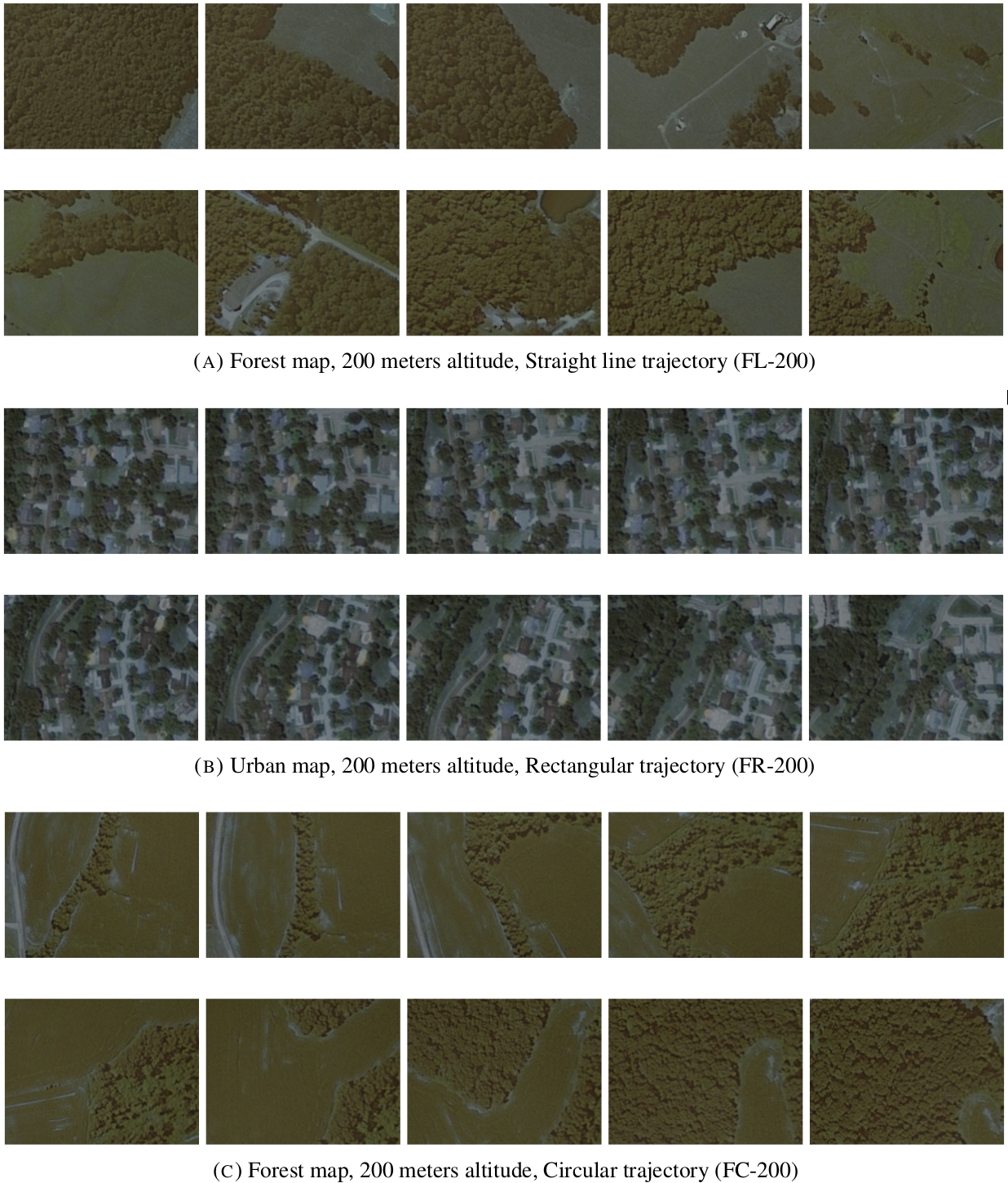}
\caption{Sample image sequences from the dataset. }
\label{fig:sample_images}
\end{figure*}

Dataset consists of 12 archives for each of the scenarios planned in section \ref{sec:scenarios}. Each archive contains jpeg encoded images, flight plans from QGroundControl software and a CSV file containing image metadata. Table \ref{tab:metadata_details} presents a detailed list and explanation of fields available in the CSV file. The dataset contains a total of 113474 images, alongside with geographical coordinates and UAV pose at the moment of image capture. The total distance covered by flight is over 33 kilometers, the statistics of each flight is available in table \ref{tab:flight_stats}. All images are captured at 640x480 resolution. Images are rectified and do not contain camera distortion, so camera calibration is not required. Gaussian noise is added to the images, so while they are rectified, images are not identical to the original map. Sample images from the dataset are shown in figure \ref{fig:sample_images}. 

\begin{figure*}
\centering
\includegraphics[width=0.7\textwidth]{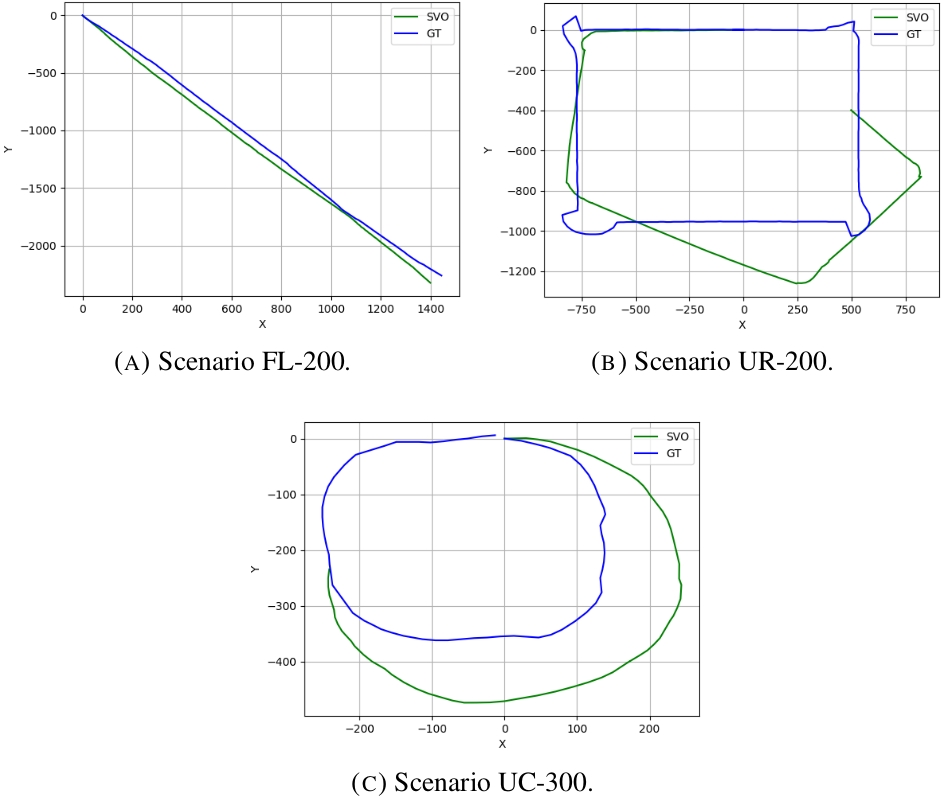}
\caption{Different flight trajectories recovered using SVO algorithm and compared against ground truth (GT).}
\label{fig:svo_traj}
\end{figure*}

To provide a baseline for comparison, the images were processed using SVO visual odometry algorithm, and the resulting coordinates are available in the metadata file. SVO was chosen since it is an open source algorithm and it achieves state-of-the-art results with small requirements in computing resources \cite{forster2014svo}. The recovered trajectories using SVO algorithm is shown in figure \ref{fig:svo_traj}. From the recovered trajectories we can see that while SVO algorithm works pretty accurate on straight line trajectory, it fails to recover rectangular and circular trajectories. The problem is that SVO fails to accurately calculate trajectory while the plane is performing turn maneuver, which causes pure rotational movement which is known to cause problems or SVO algorithm.

\newcommand{\myTrajScale}{0.215}


\newcommand{\CommaPunct}{,\space}
\begin{table*}
\caption{Metadata fields available for each image in the dataset.\label{tab:metadata_details}}
\begin{tabularx}{\textwidth}{lXl}
\toprule
\textbf{Field} & \textbf{Description} & \textbf{Unit}\\\midrule
\begin{filecontents*}{fields_short.csv}
Field,Description,Measurement unit
Filename,Name of the image file,-
Latitude,Latitude of the aircraft position ,degrees
Longitude,Longitude of the aircraft position,degrees
RelativeAltitude, Altitude from the earth’s surface,meters
ImuX, IMU orientation quaternion\CommaPunct X component,-
ImuY, IMU orientation quaternion\CommaPunct Y component,-
ImuZ, IMU orientation quaternion\CommaPunct Z component,-
ImuW, IMU orientation quaternion\CommaPunct W component,-
PoseX,GT position relative to flight start location\CommaPunct X component,meters
PoseY,GT position\CommaPunct Y component,meters
PoseZ,GT position\CommaPunct Z component,meters
OrientationX,GT orientation\CommaPunct relative to world coordinate frame\CommaPunct X component,-
OrientationY,GT orientation\CommaPunct Y component,-
OrientationZ,GT orientation\CommaPunct Z component,-
OrientationW,GT orientation\CommaPunct W component,-
MapX,Image center location on map\CommaPunct X axis,pixels
MapY,Image center location on map\CommaPunct Y axis,pixels
SvoX,Pose estimate\CommaPunct calculated using SVO algorithm in Euclidean space\CommaPunct X component,meters
SvoY,Pose estimate\CommaPunct Y component,meters
SvoZ,Pose estimate\CommaPunct Z component,meters
\end{filecontents*}
\csvreader[late after line=\\\midrule,late after last line=\\\bottomrule]
  {fields_short.csv}
  {}
  {\csvcoli & \csvcolii & \csvcoliii}
\end{tabularx}
\end{table*}

\section{Conclusion And Future Works}

Research of GPS-denied localization systems is gaining traction since UAVs are dependent on GPS signal and fail to operate without it. This paper discusses the problem of a missing dataset for the benchmark of high altitude UAV localization systems. Dataset presented in this paper contains data from 12 simulated flights totaling in 33 kilometers of flight distance, so it is by far the largest publicly available dataset of downward facing camera imagery from high altitude UAV flights. The dataset was created with as a benchmark for map-based localization system, but it also contains enough information to be used as an additional benchmark for map-less and map-building systems, such as visual odometry, optical flow, and SLAM systems.
Future works include the extension of the dataset with imagery from real-flights, include additional information from sensors, such as raw IMU readings. To improve the credibility for benchmarks, the dataset would benefit from additional imagery captured in real-flight scenarios. 

\bibliographystyle{agsm}

\bibliography{IEEEabrv,bibliography.bib}

\end{document}